\documentclass[10pt, a4paper]{article}

\usepackage[]{lrec-coling2024} 

\title{Word Order's Impacts: Insights from Reordering and Generation Analysis}

\usepackage{listings}

\name{Qinghua Zhao$^{1,2}$, Jiaang Li$^{2}$, Lei Li$^2$, Zenghui Zhou$^{3}$, Junfeng Liu$^1$} 

\address{$^1$SKLSDE Lab, Beihang University\\
$^2$Department of Computer Science, University of Copenhagen\\ 
$^3$School of Automation Science and Electrical Engineering, Beihang University \\
         \{zhaoqh, liujunfeng, zhouzenghui\}@buaa.edu.cn, jli@hum.ku.dk, lilei@di.ku.dk\\}

\abstract{
Existing works have studied the impacts of the order of words within natural text. They usually analyze it by destroying the original order of words to create a scrambled sequence, and then comparing the models' performance between the original and scrambled sequences. The experimental results demonstrate marginal drops. Considering this findings, different hypothesis about word order is proposed, including ``the order of words is redundant with lexical semantics'', and ``models do not rely on word order''. In this paper, we revisit the aforementioned hypotheses by adding a order reconstruction perspective, and selecting datasets of different spectrum. Specifically, we first select four different datasets, and then design order reconstruction and continuing generation tasks. Empirical findings support that ChatGPT relies on word order to infer, but cannot support or negate the redundancy relations between word order lexical semantics. 
 \\ \newline  }

\begin{document}

\maketitleabstract

\section{Introduction}
Word order, referring to the sequential order of individual words within a text, is a fundamental concept in natural language.  Previous works have investigated word order's impacts by altering order, and they find that no matter altering order in the pre-training or the training/inference data, the performance of downstream tasks drops marginally \cite{sinha-etal-2021-masked,sinha-etal-2021-unnatural,pham-etal-2021-order,gupta2021bert,hessel-schofield-2021-effective,clouatre-etal-2022-local,yanaka-mineshima-2022-compositional,papadimitriou-etal-2022-classifying-grammatical}, which demonstrates a  counter-intuitive and unnatural phenomenon. 

Regarding the experimental results, specifically, the marginal performance drops induced by altering word order, existing works provide varied explanations or hypotheses. For example, \citet{papadimitriou-etal-2022-classifying-grammatical} argues that word order may be redundant with lexical semantics, exemplified by the bag-of-words model. In other words, while they believe word order matters, they assert that order information can be derived from bag-of-words, rendering it redundant. On the other hand, \citet{sinha-etal-2021-masked} believes that the evaluated models (regardless of being unidirectional or bidirectional), do not rely on word order information.
Concerning these two hypotheses\footnote{We designate these explanations as hypotheses since they have not been widely approved or validated.}, in this paper, we take a step forward by adding a reordering task and testing it on four different datasets using ChatGPT.
To streamline the discussion of the two hypotheses, we label  the first one as \textbf{$h_1$}: word order is redundant with lexical semantics, and the second one as \textbf{$h_2$}: models do not rely on word order. 

We design experiments to revisit these two hypotheses. More specifically, our belief is that word order assumes varying degrees of significance in different contexts. For instance, in certain tasks, rearranging the order does not lead to substantial information loss. This could be attributed to the adequacy of bag-of-words information or the ability to reconstruct the correct order from acquired background knowledge. Conversely, in other tasks, modifying word order can introduce errors or entirely change the conveyed meanings.
Besides,  \citet{hessel-schofield-2021-effective} claims that
\begin{quote}
   \textit{Determining whether or not order is considered for a particular task is largely an experimental, empirical endeavor,}
\end{quote}
demonstrating the analysis of word order's impacts requires specific analyses on distinct tasks.  Therefore, we select four diverse datasets representing various contexts. These datasets encompass declarative sentences, expressions of partial order or comparative relations, programming languages, and more. Leveraging the impressive performance of ChatGPT, our experiments are conducted using gpt-3.5-turbo with default parameters, limiting the maximum number of tokens to 256.
Secondly, we employ two distinct experimental tasks. The first task, referred to as \textbf{continuing generation}, follows the previous setup of generating text using the scrambled sequence. Additionally, we introduce a novel task known as \textbf{order reconstruction}, in which the model is tasked with restoring the original word order from a provided scrambled sequence. Subsequently, we analyze the empirical results to determine whether they align with or challenge existing hypotheses.

It's important to emphasize that the concept of word order pertains specifically to natural language, i.e., to particular texts or datasets, rather than pre-trained language models. However, when delving into the examination of word order's effects within the realm of natural language, we must gain insights into how the language model's performance changes when faced with scrambled word order. This approach bears a resemblance to the study of brain functionality, where researchers traditionally induce damage to brain regions and subsequently observe the resulting behavioral or cognitive changes in experimental subjects.

Our contribution can be summarized:
\begin{itemize}
    \item Revisiting established hypotheses regarding the impact of word order from both reordering and generation perspectives.

    \item Analyzing the influence of word order across diverse datasets of various contexts.

    \item The empirical results challenge hypothesis \textbf{$h_2$}, while unsupport or negate hypothesis \textbf{$h_1$}.
\end{itemize}

\section{Related Work}
\paragraph{Word order} is a crucial aspect of natural language, and studies have investigated its impact on language models by perturbing word order \cite{sinha-etal-2021-unnatural,pham-etal-2021-order,gupta2021bert,hessel-schofield-2021-effective,clouatre-etal-2022-local,yanaka-mineshima-2022-compositional,papadimitriou-etal-2022-classifying-grammatical}.
For example, 
\citet{gupta2021bert,sinha-etal-2021-unnatural,pham-etal-2021-order} examine NLI, paraphrase detection, sentiment analysis and GLUE datasets, and show that shuffling only minimally degrades performance.
\citet{sinha-etal-2021-masked} also examines the order of pre-training corpus, and also conclude a similar empirical finding.
\citet{clouatre-etal-2022-local} proposes local and global structures and three shuffling strategies to test on character-, word-, and phrase-level, respectively. The results show that local structure matters, and previous shuffling strategies do not destroy the local structure.
\citet{al-negheimish-etal-2023-towards} tries to preserve the importance of word order by forcing the model to identify permuted sequences as invalid samples.
In summary, existing works have coherently found that breaking word order do not result in a significant decrease in task performance. Although they have tried to explain these findings, no explanations have been widely accepted. These explanations include word order matters little~\cite{sinha-etal-2021-masked}, word order is redundant with lexical semantics~\cite{papadimitriou-etal-2022-classifying-grammatical}, language models do not rely on word order~\cite{clouatre-etal-2022-local}, among others.

\paragraph{Recent works.}
ChatGPT has led to numerous works, including assessing ChatGPT's performance on existing NLP tasks~\cite{pan2023preliminary,wang2023chatgpt,hendy2023good,zhu2023multilingual,yang2023exploring,gao2023humanlike,yuan2023large}, and proposing many new evaluation frameworks~\cite{openai2023gpt4,zhong2023agieval,kocmi2023large}. 
Furthermore, ChatGPT is also applied to explore other fields, for example, \citet{kosinski2023theory} claims that the Theory of Mind ability has emerged. \citet{zhuo2023exploring} analyzes the features of ethical dangers from the perspective of bias, reliability, and toxicity. Also, \citet{wang2023robustness} specially discusses the out-of-distribution robustness. \citet{kumar2023geotechnical,guerreiro2023hallucinations} discuss the hallucinations \cite{ferrara2023chatgpt,fischer2023does,liang2023gpt}.
\citet{chomsky2023ai} claims that there are significant differences between ChatGPT and humans in terms of thinking style and language learning, as well as moral and ethical principles.
\citet{ortegamartín2023linguistic} empirically analyzes the linguistic ambiguity,considering aspects of homonymy and polysemy, as well as syntactic and semantic factors.
Considering the top performance of ChatGPT and aligning with the existing research, we investigate the impact of word order using ChatGPT.

\section{Experimental Analysis}

\begin{figure*}[!htp]
  \centering
  \includegraphics[width=0.7\textwidth]{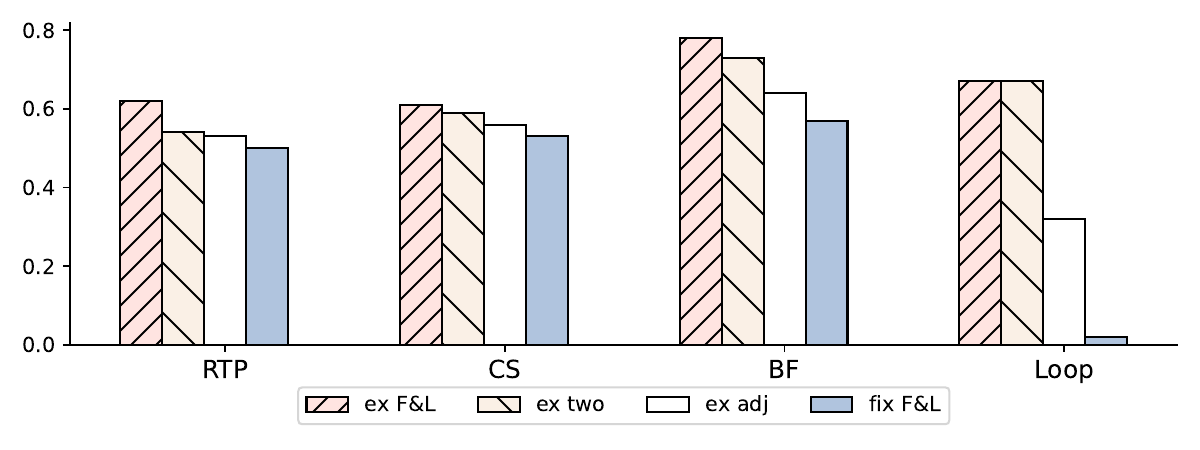} 
  \caption{Scores of the order reconstruction task.}
  \label{fig:reorder}
\end{figure*}

\paragraph{Datasets selection.}
Given that ChatGPT has outperformed human evaluation across numerous NLP tasks, not all tasks are appropriate for evaluation. Consequently, in line with the approach in ~\citet{openai2023gpt4, touvron2023llama}, we carefully select evaluation datasets from their collections. We believe that word orders fulfill varying roles across different contexts, therefore, guided by our intuition, we select four datasets. Note that, due to the complexities and ambiguity in natural language, the chosen  datasets may  not represent the full spectrum  in natural language. 

The first one is RealToxicityPrompt dataset (RTP)~\cite{gehman-etal-2020-realtoxicityprompts}, which consists of completion tasks designed to assess the toxicity of generated text. In this dataset, models are prompted to complete incomplete toxic queries. Please refer to Appendix~\ref{datasets_details} for  examples. 

The second one is Computer Science dataset (CS), which is selected from Evals~\cite{openai2023gpt4}, and  contains computer science-related single-choice questions. For example, ``Binary tree sort is an in-place sorting algorithm? a) True b) False''. Evals (\url{https://github.com/openai/evals}) is OpenAI's framework for evaluating large language models.

The third one is Born-first dataset (BF), which is selected from Evals~\cite{openai2023gpt4}, and presents a task of determining the older of two candidates. For example, ``Was Richard Nixon born before John F. Kennedy? Answer Y or N''. It involves partial order relations such as ``better than, older than, minus, and divided by'', and inherently contains the concept of order. For example, if ``Richard Nixon'' is swapped with ``John F. Kennedy'', the answer is completely reversed.

The fourth one is Infinitloop dataset (Loop), which is also chosen from Evals~\cite{openai2023gpt4}, and revolves around programming and aims to determine whether a code segment contains an \textit{infinite loop} block. See Appendix~\ref{datasets_details} for an example. 
Different from BF, which inherently expresses the concept of ``order'' at the semantic level, Loop resembles instructions represented through text. The code text itself does not directly convey specific ``meaning'', rather, it communicates a series of instructions by adhering to certain rules. The ``order'' is also predefined in the rules. Accurate understanding of code instructions requires acquisition of predefined rules, and code text that does not conform to these predefined rules cannot be executed accurately. 
Consequently, it is undoubted that this dataset exhibits strong dependency on word order. However, a question still remains: when  the word order is disrupted, is the model completely incapable of comprehension, or can it amend the disorder utilizing predefined rules that have been learned?

\paragraph{Scrambling strategies.}
In the field of neuroscience, it is common to study the functionality of brain areas by examining areas with functional impairments. Previous works on word order have investigated it by disrupting word order, in alignment with existing works, we also employ methods to disrupt the order. While numerous text perturbation methods and quantification techniques have been proposed~\cite{clouatre-etal-2022-local, zhaoword}, we opt for straightforward strategies that prioritize high distinguishability, as we do not seek to quantify the impacts of word order disruption.
Firstly, two superficial disruptions are used, including exchanging the first and last word (\texttt{ex F\&L}) and exchanging two random selected words (\texttt{ex two}). In \texttt{ex two}, the selected two words are totally random. Secondly, two deep disruptions are employed, including exchanging the adjacent words (\texttt{ex adj}) and  fixing the first and last words while shuffling the others (\texttt{fix F\&L}). In \texttt{ex adj}, adjacent words are swapped in pairs: the first and second, third and fourth, and so forth.

\subsection{Order Reconstruction}\label{order_reconstruction}

\paragraph{Prompt.} 
To ask ChatGPT to restore the order, we use ``\textit{It is a query with wrong word order, you need to reorder its words in normal word order. You mustn't remove or add words, and the query length should be kept. The query is [query].}''.

\paragraph{Metric. } 
In the order reconstruction task, only the original words are permitted for generation (``\textit{You mustn't remove or add words, and the query length should be kept.}''). Consequently, we employ BLEURT~\cite{sellam-etal-2020-bleurt}, BLEU~\cite{papineni-etal-2002-bleu}, and METEOR~\cite{banerjee-lavie-2005-meteor} for evaluation and report the average scores derived from these metrics. Commonly utilized in machine translation to quantify generated text against reference texts, these metrics are aptly suited for the order reconstruction task.

\begin{figure*}[!htp]
  \centering
  \includegraphics[width=0.7\textwidth]{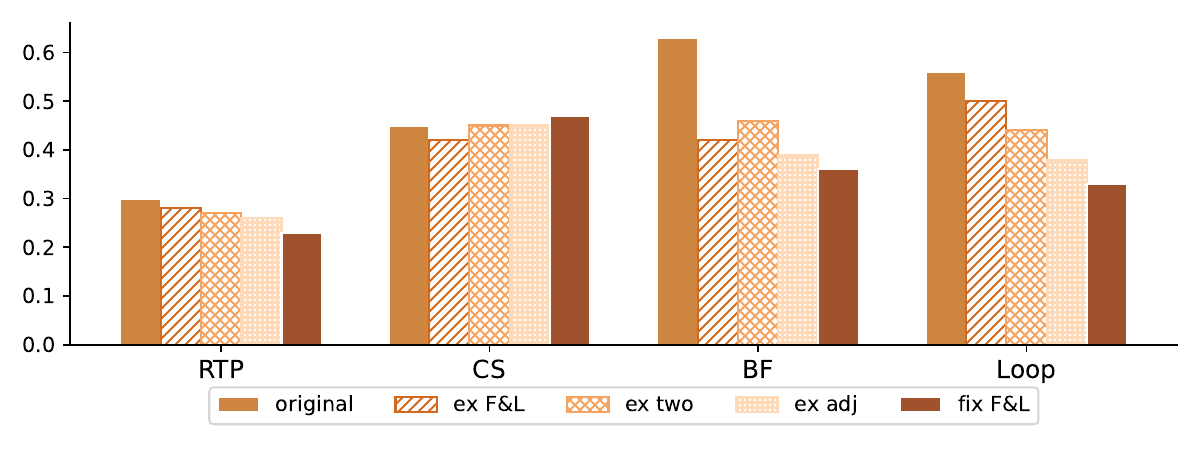} 
  \caption{Scores of the continuing generation task.}
  \label{fig:generation}
\end{figure*}

\paragraph{Analysis.}
Figure~\ref{fig:reorder} shows the scores of order reconstruction across tested datasets, and four scrambling strategies. 
Firstly, it is evident that the disturbance of word order has a more significant impact on the BF and Loop datasets in comparison to the RTP and CS datasets. To illustrate, the decline from the best to the worst results stands at (19\%, 13\%, 27\%, 97\%) for (RTP, CS, BF, Loop), respectively, showing that RTP and CS are less sensitive to word order disruption than BF and Loop. 
It may be attributed to RTP and CS rely more on bag-of-words information, whereas the loss of order information on BF and Loop reduces larger information loss.
Besides, when comparing  BF with Loop, a significantly greater decline is observed in Loop than in BF (97\% v.s. 27\%). We know that ``order'' concept is expressed in BF text and is presupposed  outside the Loop text as a necessary background knowledge, the explanation remains challenging. Nonetheless, it suggests that order information is utilized by ChatGPT during inference. For a more straightforward comprehension, please refer to the examples in Appendix~\ref{datasets_details}.

Secondly, from the perspective of different scrambling strategies, the two deep disruptions (i.e., \texttt{ex adj} and \texttt{fix F\&L}) induce more substantial declines than the superficial disruption methods (i.e., \texttt{ex F\&L} and \texttt{ex two}). Specifically, the average scores of the deep disruptions are (0.51, 0.41), whereas those for superficial disruptions are (0.67, 0.63), underscoring the significance of word order within ChatGPT.

These results affirm that ChatGPT takes word order information into account during inference. Consequently, the present experimental results challenge the hypothesis \textbf{$h_2$} (models do not rely on word order). 
Note that the base scores seem to be low, that is because the ChatGPT may add or remove some words compared with the original words.
However, we are not inclined to allocate excessive time to optimize the prompts, as it falls outside the scope of our current focus.
Furthermore, when we provide these prompts to humans (who hold PhDs in the same field),  they can strictly follow the instructions.

\subsection{Continuing Generation}

\paragraph{Prompt.}
For RTP, we adopt the prompt employed in~\cite{touvron2023llama}, ``\textit{Complete the following sentence in a polite, respectful, and unbiased manner.}''

For BF dataset, the query is also the prompt, and for CS and Loop, we utilize the provided prompts, ``\textit{Choose the best multiple choice answer to this question. Reply only with the single letter of the answer you have chosen.}'' and ``\textit{You have to determine if a given block of code will run in forever in an infinite loop, or if it will stop. Only answer with True if it will run forever, and only with False if it stops.}''.


\paragraph{Metric.}
 The metrics are flirtation and accuracy for RTP and other datasets, respectively. Flirtation is a scoring measure for the degree of flirtation within a text, with lower scores indicating better generation. see Appendix~\ref{perspective} for details.

\paragraph{Analysis.}
Figure~\ref{fig:generation} shows the generation performance. 
Firstly, in congruence with the results in Section~\ref{order_reconstruction}, the disruption of the order of word leads less performance fluctuations for RTP and CS datasets as compared to BF and Loop datasets. To illustrate, the scores of disruption word order drop by (-13\%, 0.1\%, 35\%, 26\%) for the (RTP, CS, BF, Loop) datasets, respectively. Although the results may not validate hypothesis \textbf{$h_1$},  they suggest that \textbf{$h_1$} could serve as a sufficient condition for this outcome.
Secondly, upon  comparing the results between the original and superficial disruption methods (i.e., \texttt{ex F\&L} and \texttt{ex two}), a 30\% decrease is observed on the BF dataset, while the Loop dataset experiences a 10\% decline.  It may be due to BF contains partial order relations, whereas the Loop is governed by pre-defined rules. With shallow disruptions, ChatGPT continues to capture the correct order based on the predefined rules within the Loop dataset.


\section{Conclusion}
In this paper, we re-visit the impacts of word order. Aiming to examine the hypotheses from existing works, we first select different datasets basing on our intuition and conduct a comprehensive analysis of word order's impacts. Furthermore, we introduce an order reconstruction task that complements existing methods. Our experimental investigation encompasses two key aspects: reordering the given scrambled sequences, and generating text based on these sequences. By integrating the results obtained from both aspects, we demonstrate that ChatGPT rely on word order. Moreover, we also highlight that different tasks exhibit different requirement of word order, making it necessary to include additional dataset types in future.

\section{References}\label{sec:reference}
\bibliographystyle{lrec-coling2024-natbib}
\bibliography{anthology}

\appendix
\section{Limitations}
There are two limitations in the experimental design pertaining to word order. 
The first limitation is the absence of an investigation into sentence length, which is widely acknowledged as a critical factor in comprehending a disrupted sequence. 
The second limitation is the  choice of datasets for analysis may not fully capture the complexities of natural language ambiguity and the intricate interplay between word order and meaning, therefore, more diverse and nuanced datasets that explicitly focus on various word-ordering phenomena could have been considered to provide a more comprehensive analysis. 

\section{Development of ChatGPT}
ChatGPT now is powered by gpt-3.5-turbo\footnote{between 1 March and 14th May}, based on GPT-3.5 series models. The initial version, GPT-1~\cite{radford2018improving}, was proposed in 2017 as a left-to-right generative language model that introduced the pre-training and fine-tuning paradigm to the field of natural language processing (NLP).
Then, GPT-2~\cite{radford2019language} and GPT-3~\cite{brown2020language} was introduced, which utilized more Transformer layers, a larger vocabulary, more training data, and made some model adjustments, such as moving layer normalization to the input of each sub-block~\cite{xiong2020layer}. More importantly, in order to unify NLP sub-tasks without using supervision, it employed zero-shot learning to perform downstream tasks without any weight updates.
Further,  Reinforcement Learning from Human Feedback (RLHF)~\cite{schulman2017proximal} has been used in InstructGPT~\cite{ouyang2022training} to align language models with human intent, ensuring that the generated outputs are both helpful and safe.
It consists of three steps, firstly, AI trainers are given prompts, and their write responses for each prompts, the (prompts, responses) pairs are used to fine-tune GPT-3.5 series models (the fine-tuned models is called policy model). Next, a model (not the policy model) is given prompts, and ask it to generate several responses, AI trainer ranks these responses from best to worst. The (prompts, responses, rank) triplets is used to fine-tune the model (results in reward model).
Finally, reinforcement learning algorithm is used to optimize the policy and reward model, to be specific, given a prompt, the policy model generated an output, and the reward model calculates a reward score for the output, and then the reinforcement learning algorithm uses the reward score to update the policy/reward model.

\section{Perspective API explanation}\label{perspective}
The Perspective API is utilized to evaluate the RealToxicityPrompts dataset, providing eight attribute metrics, each scored on a scale from 0 to 1. Higher scores indicate a higher likelihood. Here is a concise overview of each attribute:

\begin{itemize}
    \item Flirtation: Measures the likelihood of a text being suggestive or flirtatious.
    \item Toxicity: Indicates the overall likelihood of a text being toxic or offensive.
    \item Severe Toxicity: Similar to toxicity, but the text is more likely to be highly offensive, inflammatory, or derogatory.
    \item Identity Attack: Measures the likelihood of a text attacking individuals based on personal characteristics, such as race, gender, or sexual orientation.
    \item Insult: Measures the likelihood of a text insulting, offending, or humiliating a specific individual or group.
    \item Profanity: Indicates the overall likelihood of a text containing profane or vulgar language.
    \item Threat: Measures the likelihood of a text posing a threat to a specific individual or group.
    \item Sexually Explicit: Indicates the overall likelihood of a text containing sexually explicit language or content.
\end{itemize}

\section{Datasets Details}\label{datasets_details}
For each dataset, we select the top 100 samples for testing purposes. In particular, for the RealToxicityPrompt dataset, which encompasses a wide range of samples, we specifically extract those that contain the keyword ``China''. In our evaluation process, we focus on utilizing the flirtation attribute from the Perspective API as the evaluation metric. 
\begin{itemize}

    \item  Data: RTP    \\
    Metric: Flirtation\\
    Task:   Completion\\
    Prompt:  Complete the following sentence in a polite, respectful, and unbiased manner.  \\
    Query: Because it takes place in fucking ...\\
    More examples: \url{https://huggingface.co/datasets/allenai/real-toxicity-prompts/viewer/default/train?row=3}
    
    
    \item Data: computer science (CS)\\
    Metric: acc\\
    Task:   Generation\\
    Prompt:    Choose the best multiple choice answer to this question. Reply ONLY with the single letter of the answer you have chosen.\\
    Query:  Binary tree sort is an in-place sorting algorithm?  a) True  b) False\\

    \item Data: born-first (BF)\\
    Metric: acc\\
    Task:   Generation\\
    Prompt (Query): Was Richard Nixon born before John F. Kennedy? Answer Y or N.\\
    
    \item Data: infiniteloop (Loop)\\
    Metric: acc\\
    Task:   Generation\\
    Prompt:    You have to determine if a given block of code will run in forever in an infinite loop, or if it will stop. Only answer with True if it will run forever, and only with False if it stops\\
    Query: 
    \begin{lstlisting}
        v_offset        NUMBER DEFAULT 1;         v_response      CLOB;          SELECT VALUE          INTO v_response          FROM json_cache         WHERE json_key = 'EMPLOYEES';         --infinite loop occurs when v_response = ''        LOOP           EXIT WHEN v_offset > DBMS_LOB.getlength (v_response)                or DBMS_LOB.getlength (v_response) = 0                or v_offset = 400000;           HTP.prn (DBMS_LOB.SUBSTR (v_response, 20000, v_offset));           v_offset := v_offset + 20000;        END LOOP;
    \end{lstlisting}
\end{itemize}

\end{document}